\documentclass[letterpaper, 10 pt, conference]{ieeeconf}  % Comment this line out if you need a4paper

\IEEEoverridecommandlockouts                              % This command is only needed if
\usepackage{times} % assumes new font selection scheme installed
\usepackage{amsmath} % assumes amsmath package installed
\usepackage{amssymb,amsfonts}  % assumes amsmath package installed
\usepackage[sort,compress]{cite}
\usepackage{graphicx,balance}
\usepackage{booktabs}
\usepackage{mathrsfs}
\usepackage{subcaption}
\usepackage{tikz}
\let\labelindent\relax  % issue with enumitem
\usepackage{enumitem}  % fancy enumerations
\usepackage{algorithm}
\usepackage{algorithmicx}
\usepackage{algpseudocode}

\definecolor{DarkGreen}{rgb}{0,0.5,0}
\definecolor{DarkRed}{rgb}{0.75,0,0}

\usepackage{freytex}
\renewcommand{\C}{\mc{C}}  % candidate set
\newcommand{\CA}{{{}^A\C}}
\newcommand{\CB}{{{}^B\C}}
\newcommand{\Cgraph}{\mc{G}_c}  % correspondence graph
\newcommand{\Cliques}{\mathfrak{C}(\Cgraph)}  % set of cliques
\newcommand{\Vertices}{\mc{V}(\Cgraph)}  % set of vertices
\newcommand{\dchi}[1]{d_{\chi^2,#1}}  % chi^2 threshold

\newcommand{\dGC}{d_{\text{GC}}}
\newcommand{\cov}[1][{}]{{\mx[#1]{\Sigma}}}  % covariance matrix

\newcommand{\subscript}[2]{$#1_#2$}  % for letter-indexed enumeration

\graphicspath{{figures/}}

\title{\LARGE \bf
Efficient Constellation-Based Map-Merging for Semantic SLAM
}

\author{Kristoffer M.\ Frey$^{1}$, Ted J.\ Steiner$^{2}$, and Jonathan P.\ How$^{1}$% <-this % stops a space
\thanks{Supported by the Defense Advanced Research Projects Agency (DARPA) as part of the Fast Lightweight Autonomy (FLA) program, HR0011-15-C-0110.
        Views expressed here are those of the authors, and do not reflect the official views or policies of the Dept.\ of Defense or the U.S. Government.}% <-this % stops a space
\thanks{$^{1}$K.\ Frey and J.\ How are with the Department of Aeronautics and Astronautics, MIT,
        Cambridge, MA 02139, USA.
        {\tt \{kfrey,jhow\}@mit.edu}}%
\thanks{$^{2}$T.\ Steiner is a Senior Member of the Technical Staff at Draper, Cambridge, MA 02139, USA.
        {\tt tsteiner@draper.com}}%
}

\begin{document}

\maketitle
\thispagestyle{empty}
\pagestyle{empty}

%%%%%%%%%%%%%%%%%%%%%%%%%%%%%%%%%%%%%%%%%%%%%%%%%%%%%%%%%%%%%%%%%%%%%%%%%%%%%%%%
\begin{abstract}
  Data association in SLAM is fundamentally challenging, and handling ambiguity well is crucial to achieve robust operation in real-world environments.
  When ambiguous measurements arise, conservatism often mandates that the measurement is discarded or a new landmark is initialized rather than risking an incorrect association.
  To address the inevitable ``duplicate'' landmarks that arise, we present an efficient map-merging framework to detect duplicate \emph{constellations} of landmarks, providing a high-confidence loop-closure mechanism well-suited for object-level SLAM.
  This approach uses an incrementally-computable approximation of landmark uncertainty that only depends on local information in the SLAM graph, avoiding expensive recovery of the full system covariance matrix.
  This enables a search based on geometric consistency (GC) (rather than full joint compatibility (JC)) that inexpensively reduces the search space to a handful of ``best'' hypotheses.
  Furthermore, we reformulate the commonly-used interpretation tree to allow for more efficient integration of clique-based pairwise compatibility, accelerating the branch-and-bound max-cardinality search.
  Our method is demonstrated to match the performance of full JC methods at significantly-reduced computational cost, facilitating robust object-based loop-closure over large SLAM problems.
\end{abstract}

%%%%%%%%%%%%%%%%%%%%%%%%%%%%%%%%%%%%%%%%%%%%%%%%%%%%%%%%%%%%%%%%%%%%%%%%%%%%%%%%
%%% Introduction
\section{INTRODUCTION}
The rise of single-shot object detectors \cite{redmon2016you,liu2016ssd,ren2015faster} has led to interest in extensions of classic SLAM algorithms to include semantically-meaningful landmarks.
For mobile robots operating in the real world, the ability to detect and localize semantic objects such as cars or street signs in the environment is vital for safe and effective behavior.
Besides the benefits for motion planning, the inclusion of these semantic landmarks in the SLAM factor-graph also provides opportunities to improve data association and loop-closure, fundamental challenges in SLAM \cite{cadena2016past}.
Compared to generic point features, semantic landmarks represent whole ``objects'' in the world, making them highly distinctive.
Furthermore, modern object detectors \cite{redmon2016you,liu2016ssd} are more robust to viewpoint and lighting variation than generic image-space descriptors such as SURF \cite{bay2006surf}.
Loop-closure is crucial for autonomous systems operating without the aid of GPS or other localization infrastructure, as any pure-odometry solution will drift as error accrues over time.
This drift can make data association highly ambiguous, especially under nonlinear measurement-modalities, such as vision \cite{nicholson2018quadric}.
Because a single incorrect association can be catastrophic, such systems in practice must use conservative recognition thresholds, choosing instead to initialize a new landmark whenever current measurements are poorly explained by the current set of landmark estimates.
This conservatism is always ``safe'', in the sense that the attribution of data to a new ``duplicate'' landmark will not cause estimator divergence, but it does come at the cost of increased model complexity and the loss of a valuable loop-closure constraint.
Given that some level of front-end conservatism is unavoidable for robustness on real-world data, we propose a method for identifying and merging these duplicates in the SLAM back-end.

\begin{figure}[t]
  \centerline{\includegraphics[width=0.47\textwidth]{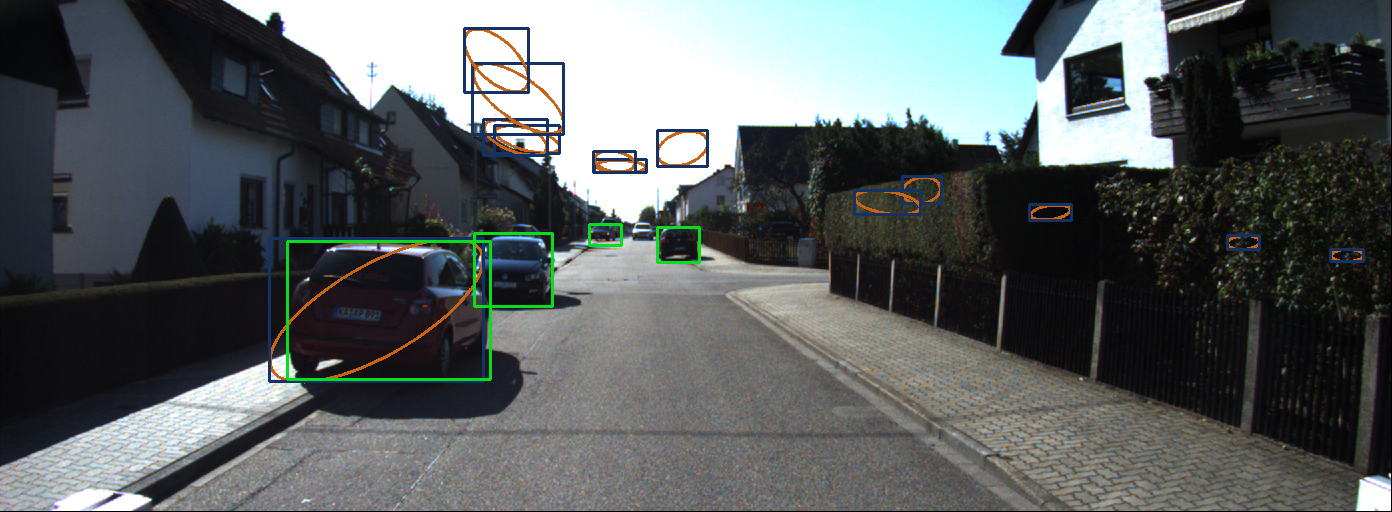}}
  \caption[]{
    SLAM algorithms are often presented with difficult data association tasks, especially after returning from long loops.
    Measurements generated by an object detector are shown as green bounding boxes, and the projections of estimated ellipsoidal landmarks are shown in orange, with predicted bounding boxes in blue.
    Rather than (potentially incorrectly) attempting to associate current measurements to existing landmarks, it is always safe to initialize a new landmark (e.g.\ car on left).
    Constellations of these duplicates can be merged together in a delayed fashion, providing a form of ``lazy'' data association.
  }
  \label{fig:kitti_diff_association}
  \vspace{-0.5cm}
\end{figure}

This paper presents a robust and efficient framework for efficient map-merging that is well-suited for semantic object-based SLAM.
In order to achieve high precision in detecting duplicate landmarks, our approach identifies maximum-cardinality \emph{constellations} of landmarks, sharing many similarities with joint gating techniques \cite{neira2001data,neira2003linear,li2012ipjc}.
However, most existing methods assume constant-time access to the SLAM covariance matrix, which in recent nonlinear approaches to SLAM \cite{dellaert2006square,kaess2011isam2} is not maintained explicitly and is expensive to recover.

To accomplish this, we define a geometric compatibility (GC) cost between two candidate constellations, capturing the most important correlations between landmarks but avoiding the reliance on global uncertainty information represented by the full covariance matrix.
Furthermore, we introduce a conservative approximation of \emph{local} landmark uncertainty that is incrementally-computable and facilitates efficient (linear-time) evaluation of the GC.
Using GC-based search to eliminate the vast majority of hypotheses, we can restrict estimation of the JC (and requisite global uncertainty information) to only the most likely candidates (as an optional verification step).

As a secondary contribution, we propose a reformulation of the JC search as a set inclusion problem over a correspondence graph.
In contrast to the traditional interpretation tree \cite{grimson1990object,neira2001data,li2012ipjc}, we employ a more flexible binary search tree (BST) that is naturally constrained to cliques on the correspondence graph (which encode satisfaction of unary and binary constraints).
This facilitates a stronger bounding for the branch-and-bound maximization, resulting in a significantly-accelerated optimization.

Our GC metric is verified on synthetic data with varying levels of noise, demonstrating desirable statistical properties in spite of nonlinear observations and significant estimate drift.
Our approach achieves comparable performance to JC search (which requires global covariance information) at a fraction of the computational cost.

%%%%%%%%%%%%%%%%%%%%%%%%%%%%%%%%%%%%%%%%%%%%%%%%%%%%%%%%%%%%%%%%%%%%%%%%%%%%%%%%
%%% Related Work
\section{Related Work}

The map merging and data association in this paper touches many subfields of SLAM (and robotics in general).
As a solution to the loop-closure and relocalization problem, it provides an alternative to direct image-based localization methods, such as \cite{newman2005slam,paul2010fab,williams2011automatic}.
However, such systems require a rich database of ``places'' (i.e., keyframes) to localize against, and are sensitive to variations in viewpoint, lighting, or scene change.
Recently, \cite{garg2018lost} achieved good robustness to extreme viewpoint and appearance variation by leveraging semantic information (similar in spirit to our approach) via per-pixel semantic classification.

Data association in SLAM is a  well-explored problem, with solutions in applications ranging from acoustic sensing \cite{huang2015towards,fallon2013relocating}, monocular vision \cite{clemente2007mapping}, and LIDAR \cite{guivant2001optimization}.
Traditionally, the search for jointly-compatible hypotheses has been approached as a search over the interpretation tree \cite{grimson1990object,hahnel2005towards} or a max-clique problem over a correspondence graph \cite{bailey2000data,san2013robust}.
The probabilistic Joint Compatibility (JC) metric introduced by \cite{neira2001data} has been widely used as the de facto standard in filtering approaches (in which the full covariance matrix is readily available).
A number of hybrid approaches \cite{neira2003linear} have been proposed, generally leveraging the correspondence graph to generate pairwise-compatible \cite{grimson1990object,lim1999mobile} hypotheses and using JC to verify them \cite{paz2008divide}.
In the case of feature cloud matching, which applies directly to sensor modalities such as LIDAR, \cite{li2012ipjc,shen2016fast} leverage the specific independence structure to provide linear and constant-time incrementalized evaluations of JC.
In contrast to the traditional gating formulation, the equivalent posterior form of the test is exploited in \cite{li2014fast} that conveniently decouples different components of the error, allowing sequential computation.

Various other (non-gating) approaches to joint data association exist, including RANSAC \cite{fischler1981random}, voting schemes \cite{paz2005global}, and explicit maximum-likelihood search \cite{atanasov2014semantic}.
Alongside these, several robust back-end approaches exist to identify and disable outlier measurements, based on residual gating \cite{graham2015robust}, linear programming \cite{carlone2014selecting}, Expectation-Maximization \cite{bowman2017probabilistic}, and explicit integer optimization \cite{sunderhauf2012switchable}.
Our approach is complementary to these, in that it adds a ``second-stage'' data-association, generating high-confidence candidate loop-closures which can be further verified by a robust back-end.

More along the lines of this work, \cite{mu2016slam} merge ``duplicate'' landmarks based on a semantic-aware clustering algorithm.
However, the clustering formulation is somewhat limited, as data associations (i.e.\ merge decisions) are made individually rather than jointly.

%%%%%%%%%%%%%%%%%%%%%%%%%%%%%%%%%%%%%%%%%%%%%%%%%%%%%%%%%%%%%%%%%%%%%%%%%%%%%%%%
%%% Preliminaries
\section{Preliminaries}

The goal of SLAM is to estimate a set of poses $\mc{T} = \{\pose_i\}$ and landmarks $\mc{L} = \{L_j\}$ given a set of noisy sensor measurements $\{z_k\}$.
Each pose is a rigid-body transform $\pose_i = (\rot_i, \vc{t}_i)$, with $\rot_i \in \SO(3)$ and $\vc{t}_i \in \R^3$ the rotation and translation components, respectively.
Notationally, we will use leading superscripts when referring with respect to a specific coordinate system, e.g.\ $\pose[w]_i$ refers to the pose with respect to the global frame $w$.
Because we are interested particularly in the \emph{semantic} variant of SLAM, each landmark $L_j$ may represent not just a point position $\vc{p}_j \in \R^3$ in space but also a class label $c_j$ and appearance descriptor $\Omega_j$.
In the authors' experience, state-of-the-art object detectors \cite{redmon2016you,liu2016ssd} achieve good classification accuracy, and to simplify the discussion in this paper we assume class labels $\{ c_j \}$ are known accurately and can be leveraged as a hard constraint when identifying duplicate landmarks (see Sec.\ \ref{ss:pairwise_compat}).  % could drop ref for space

In particular, we focus on ``relative'' SLAM problems, in which globally-referenced measurements (i.e.\ from GPS) are unavailable, and the estimation problem is only defined up to an arbitrary navigation frame $w$.
Specifically, this means that observation factors $\phi_{\text{obs}}$ and odometry factors $\phi_{\text{odom}}$ are functions that can be expressed
\begin{align}
  \phi_{\text{obs}}(\pose[w]_i, \vc[w]{p}_{j})  &= f\big( \rot[w]_i^T ( \vc[w]{p}_{j} - \vc[w]{t}_i ) \big)  \label{eq:rel_factors}  \\
  \phi_{\text{odom}}(\pose[w]_i, \pose[w]_{i+1}) &= g\big( \rot[w]_i^T ( \vc[w]{t}_{i+1} - \vc[w]{t}_i ), \rot[w]_i^T \rot[w]_{i+1} \big)  \nonumber
\end{align}
The resultant probabilistic estimation problem can be visualized as a factor graph \cite{dellaert2006square} with variables nodes representing the quantities to be estimated $(\mc{T}, \mc{L})$, and factor nodes representing the measurement models and sensor data relating them.

SLAM problems can often be quite large, involving hundreds of poses and thousands of landmarks.
Modern smoothing approaches \cite{dellaert2006square,kaess2011isam2} capitalize on the natural sparsity of such systems to perform efficient non-linear inference.
This is accomplished in part by avoiding computation of the fully dense covariance matrix $\cov[w]$, a contrast to traditional filtering approaches.
$\cov[w]$ can be recovered at any time, but requires a large (and computationally expensive) $n \times n$ matrix inversion (where $n$ is scalar dimension of the system).

While recovering the full $\cov[w]$ is computationally intractable for many SLAM applications, the minimal ``query'' required to evaluate a constellation match hypothesis involves only a limited sub-block.
Computing these marginal sub-blocks in general requires a partial Gaussian elimination over the factor graph, followed by a relatively small matrix inversion, and efficient algorithms have been proposed \cite{kaess2009covariance}.
If the number and size of such queries is limited to only the most promising hypotheses, this computation can be affordable in practice.

\subsection{Delayed Data Association as Map-Merging}
  Data association can be viewed as the problem of finding the optimal mapping between a set of $m$ measurements and $n$ estimated landmarks, where it is possible for some measurements to be spurious, or to arise from previously-unseen ``new'' landmarks.
  In our context of map-merging, we attempt to find correspondences from \emph{landmarks} to landmarks.

  Notationally, we consider \emph{candidate matches} to be ordered pairs $s = (a,b)$ where $a \neq b \in \{1, 2, \ldots, n\}$ are the indices of two estimated landmarks $L_a, L_b \in \mc{L}$.
  The ordering of these indices is significant, because it implies that a candidate \emph{merge set} $\C = \{ s_k \}$ is composed of two well-defined ``constellations'' $\CA = \{a : (a,b) \in \C \}$ and $\CB = \{b : (a,b) \in \C \}$.
  For convenience, we will also assume that merge sets are ordered.
  Thus we can index into $\CA = (a_1, a_2, \ldots, a_m)$ and $\CB = (b_1, b_2, \ldots, b_m)$.

\subsection{Joint Compatibility}
  First we introduce a landmark-to-landmark variant of the joint compatibility (JC) criteria introduced by \cite{neira2001data}.
  Replacing the ``observation model'' in feature-to-landmark association, define the matching residual between any two landmarks $(L_a, L_b)$
  \begin{equation}\label{eq:residual}
    \vc{r}(s) \triangleq \vc{p}_a - \vc{p}_b
  \end{equation}

  Under the JC framework, this residual is evaluated in the \emph{global} frame, and thus statistically depends on the global-frame covariance $\cov[w]$.
  For a given hypothesis $\C = (\CA, \CB)$ of cardinality $m$, define a stacked residual $\vc[w]{r}^T = [ \vc[w]{r}(s_1)^T,  \vc[w]{r}(s_2)^T, \ldots, \vc[w]{r}(s_m)^T ]$ and corresponding covariance
  \begin{equation}\label{eq:jc_cov}
    \cov[w]_{\vc{r}} = \cov[w]_{\CA} + \cov[w]_{\CB} - \cov[w]_{\CA,\CB} - \cov[w]_{\CB,\CA}
  \end{equation}
  where $\cov[w]_{\CA}$, $\cov[w]_{\CB}$, and $\cov_{\CA,\CB}$ are the corresponding $3m \times 3m$ sub-blocks of $\cov[w]$.

  The JC criterion can be expressed
  \begin{equation}\label{eq:jc}
    \vc[w]{r}^T \cov[w]_{\vc{r}}^{-1} \vc[w]{r} < \dchi{3m}
  \end{equation}

  It should be noted that using JC as a search criteria over the combinatoric set of joint hypotheses over all $\bigO(n^2)$ candidate matches requires computing (\ref{eq:jc_cov}) and therefore the full $\cov[w]$.
  $\cov[w]$ can of course be pre-computed before starting the search, but this in general must be repeated at each time step (as loop-closure or linearization point updates can affect covariances globally).
  Furthermore, the presence of nonlinearities in the SLAM system can result in overconfidence in the linearized uncertainty estimate.
  For these reasons, we replace the JC criterion with a geometric compatibility (GC) condition that avoids this dependence on global covariance information, facilitating computationally-lightweight and accurate gating.

\subsection{Pairwise Compatibility over a Correspondence Graph}\label{ss:pairwise_compat}
  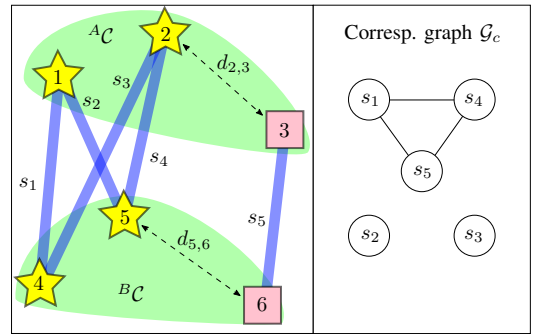
\begin{figure}[t]
    % Tikz-augmented inkscape import
    \centering
    \begin{tikzpicture}[scale=0.8, every node/.style={transform shape}]
      \node[anchor=south west,inner sep=0] (image) at (0,0)
          {\includegraphics[width=\columnwidth]{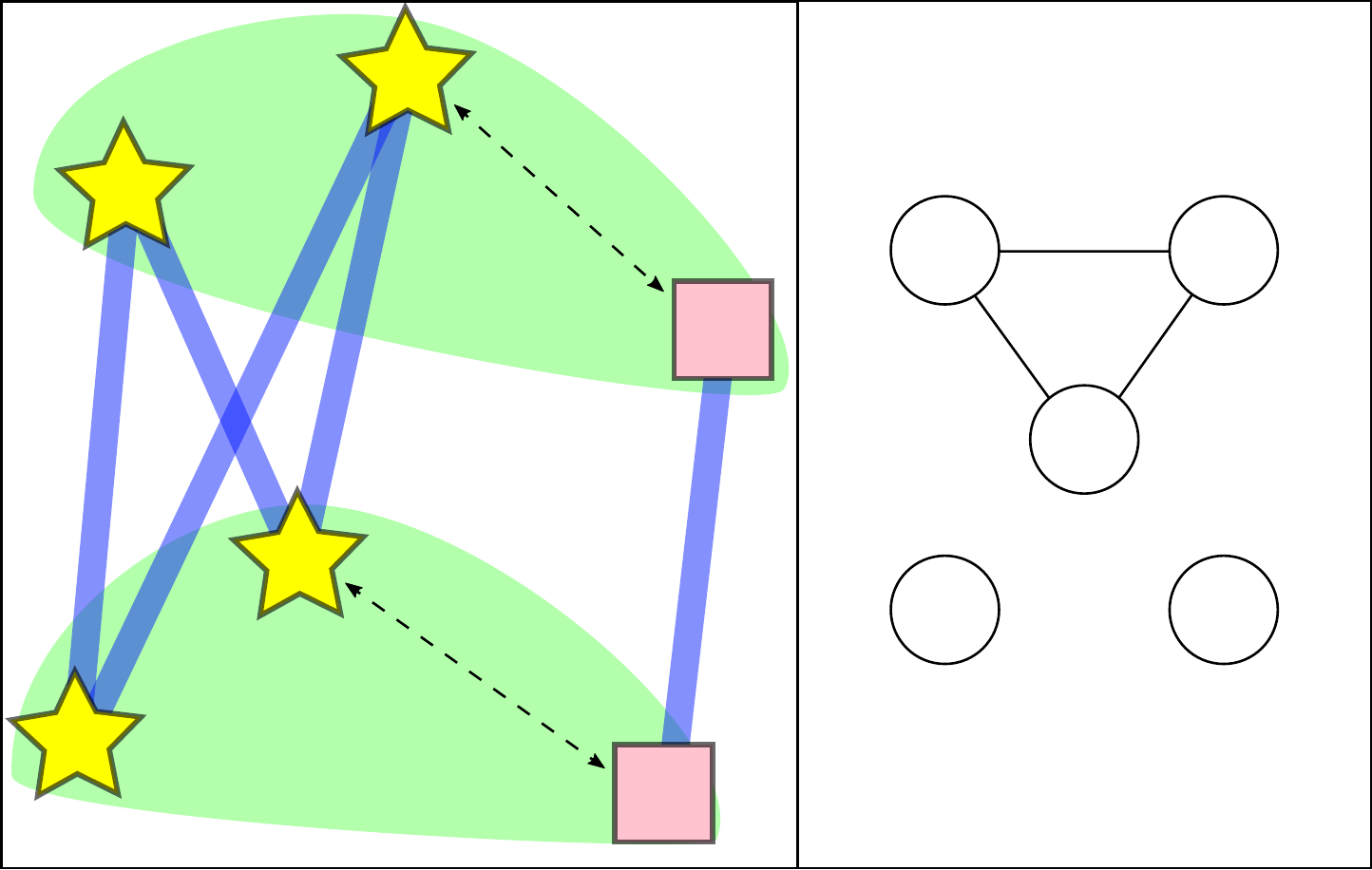}};
      % Add labels
      %\draw[help lines] (0,0) grid (8,4);

      % Landmarks
      \node at (0.79,4.27) {$1$};
      \node at (2.58,4.99) {$2$};
      \node at (4.55,3.4) {$3$};
      \node at (0.47,0.85) {$4$};
      \node at (1.9,1.95) {$5$};
      \node at (4.18,0.5) {$6$};

      % Candidate pairs
      \node at (0.30,2.5) {$s_1$};
      \node at (1.35,3.85) {$s_2$};
      \node at (1.85,4.2) {$s_3$};
      \node at (2.47,2.91) {$s_4$};
      \node at (4.1,1.9) {$s_5$};

      % Constellation labels
      \node at (1.5,5.0) {$\CA$};
      \node at (2,0.7) {$\CB$};

      % Pairwise distances
      \node at (3.71,4.48) {$d_{2,3}$};
      \node at (3.05,1.55) {$d_{5,6}$};

      % Correspondance graph
      \node at (6.85,5.0) {Corresp.\ graph $\Cgraph$};
      \node at (5.98,3.9) {$s_1$};
      \node at (7.7, 3.9) {$s_4$};
      \node at (6.85,2.7) {$s_5$};
      \node at (5.98,1.63) {$s_2$};
      \node at (7.7,1.63) {$s_3$};
    \end{tikzpicture}
    \caption{
        Drift in translation and rotation manifests as ``duplicate'' landmark constellations $\CA$ and $\CB$ (left).
        Class labels are drawn as either a star or square, and unary-compatible matches $\{s_i\}$ are shown in blue.
        A necessary condition for joint compatibility is preservation of distances, e.g.\ that $d_{2,3} \approx d_{5,6}$.
        This pairwise ``rigidity'' criterion can be formulated as a binary constraint between pairs $(s_i, s_j)$ as in \cite{grimson1990object}.
        Satisfaction of all binary constraints between candidates $(s_i, s_j)$ induces an edge in $\Cgraph$ (right), and pairwise-compatible hypotheses form cliques.
      }
    \label{fig:corr_graph}
    \vspace{-0.5cm}
  \end{figure}

  Fundamental to our approach is a tight integration between explicit tree search and clique-based compatibility.
  Given a set of unary and binary constraints on candidate matches and pairs of matches, respectively, a correspondence graph \cite{lim1999mobile,bailey2000data} can be defined.
  Candidates $s_i$ satisfying unary constraints are represented as nodes, with edges connecting pairs $(s_i, s_j)$ that satisfy binary constraints.
  Thus, cliques on this graph represent \emph{pairwise}-compatible sets of candidates (a weaker condition than joint compatibility).
  The unary and binary constraints in question are generic, and can represent geometric constraints \cite{grimson1990object,lim1999mobile}, locality \cite{neira2003linear}, appearance and class similarity, or other expert knowledge.

  If it is assumed that these constraints are in fact \emph{sufficient} for joint compatibility, the search problem can be formulated as a max-clique problem \cite{bailey2000data,lim1999mobile,san2013robust}.
  While this has the benefit of taking advantage of off-the-shelf graph-theoretic algorithms, the sufficiency assumption can be limiting.
  Nevertheless, this correspondence graph can be much more efficient than naive JC tree search, as the binary constraints effectively eliminate entire branches of the tree, and are evaluated only once for each pair $(s_i, s_j)$.
  As explained in Section \ref{s:hybrid_bb}, the correspondence graph provides a hitherto unexploited tight upper bound for max-cardinality maximization that can significantly accelerate the search.

  A number of unary and binary constraints may apply to the scenario of semantic map-merging.
  The minimal set of constraints assumed in this paper is given below.

  \noindent\textbf{Unary: } $s = (a, b)$
  \begin{enumerate}[label=(\subscript{U}{{\arabic*}})]
    \item Disjoint: $a$ not equal to $b$
    \item Class label match: $c_a = c_b$  \label{uc:class}
    \item Min separation (Sec.~\ref{ss:local_covs})  \label{uc:separation}
    %\item Conservative translation error bound (Section \ref{s:geo_constraints})  \label{uc:drift}
  \end{enumerate}

  \noindent\textbf{Binary:} $s_1 = (a_1, b_1)$, $s_2 = (a_2, b_2)$
  \begin{enumerate}[label=(\subscript{B}{{\arabic*}})]
    \item Disjoint: $a_1, b_1, a_2$ and $b_2$ are distinct indices  \label{bc:disjoint}
    \item Locality: ($a_1$, $a_2$) ``close'', ($b_1$, $b_2$) ``close'' (Sec.~\ref{ss:local_covs})  \label{bc:locality}
    %\item Preservation of distances (Section \ref{s:geo_constraints})  \label{bc:dist}
    %\item Conservative rotation error bound (Section \ref{s:geo_constraints})  \label{bc:drift}
  \end{enumerate}

\subsection{Max-Cardinality Search}
  In practice, we wish to find the largest set of correspondences that satisfies a suite of compatibility conditions.
  As an optimization, this can be formulated as a max-cardinality search over cliques $\Cliques$ in our correspondence graph $\Cgraph$.
  \begin{align}
    \max_{\C \in \Cliques}        &\qquad \lvert \C \rvert  \label{eq:jc_max} \\
    \text{subject to: } \quad     &\qquad \tt{Compatible}(\C)  \label{eq:comp_constraint}
  \end{align}
  Without the compatibility constraint (\ref{eq:comp_constraint}), this reduces to a max-clique search, as in \cite{bailey2000data}, but with it an explicit tree search is required.

  The combinatoric search over all possible assignments $\C$ has traditionally been visualized as an interpretation tree \cite{grimson1990object,neira2001data}.
  The interpretation tree has a branching factor of $n+1$ (the number of matchable landmarks plus a ``null'' match), and a depth of $m$ (the number of ``measurements'').
  The path from the root to any leaf node describes a potential joint assignment $\C$, and the structure of the tree compactly imposes the constraint that a single measurement cannot match to more than one landmark.
  At each step of the search, a partial hypothesis is evaluated, and if the constraint (\ref{eq:comp_constraint}) is not satisfied, the algorithm discontinues exploration of the corresponding branch.
  Thus, a strong compatibility metric (i.e.\ JC) is vital to efficiently prune the search.

  Though standard, the interpretation tree framework (and associated algorithms) has a significant weakness: it cannot efficiently represent the requirement that $\C \in \Cliques$.
  Though unary and binary constraints can of course be checked at each step of the tree search \cite{neira2003linear}, this is inherently inefficient because the same candidate matches $s_i, s_j$ will be tested multiple times.
  This fact has historically made max-clique and tree-search approaches largely disparate, with ``hybrid'' algorithms essentially switching from clique-based hypothesis generation to tree-based verification \cite{neira2003linear}.
  Additionally, efficient branch-and-bound maximization requires the availability of tight upper bounds.
  However, the main feasibility criteria $\C \in \Cliques$ cannot be directly ``read'' from the interpretation tree, and thus much weaker bounds based solely on tree depth are used in practice \cite[Alg.~2]{clemente2007mapping}.
  In Section \ref{s:hybrid_bb}, we reformulate the interpretation tree as a set inclusion problem over a binary-search-tree (BST), a more flexible framework allowing tighter (and therefore more efficient) branch-and-bound search.

%%%%%%%%%%%%%%%%%%%%%%%%%%%%%%%%%%%%%%%%%%%%%%%
\section{Geometric Compatibility and Local Uncertainty}
  Assume we are given two constellations $\CA$ and $\CB$ of cardinality $m > 1$ which are well-separated in the graph (i.e.\ that estimate correlations are small between them, as might be the case when a robot returns from a long loop).
  If the two constellations can be considered ``locally rigid'' (a concept which will be made more precise in the following section), the JC error can be thought of as simultaneously capturing geometric error (how well constellations ``match'' under optimal alignment) and drift error (distance in translation and rotation).
  Importantly, the geometric error is a local property (involving only the local subgraphs of $\CA$ and $\CB$ respectively) while the drift error is a global property of the posterior distribution.
  To exploit this fact, we introduce a convenient approximation of local uncertainty that decouples the geometric error from the rest of the graph.
  This decoupling allows efficient search for maximum-cardinality joint hypotheses $\C$ satisfying this GC criterion, which can then be verified globally via a JC test as an optional post-step.

  \subsection{Geometric Compatibility (GC)}\label{ss:gc}
    In order to decouple the joint compatibility between constellations $\CA = \{a_1, a_2, \ldots, a_m\}$ and $\CB = \{b_1, b_2, \ldots, b_m\}$, we consider the geometric fit given the \emph{optimal} rigid-body alignment.
    In the general case, we then have two corresponding sets of landmark estimates, $\{ \vc[A]{p}_{a_i} \}$ and $\{ \vc[B]{p}_{b_i} \}$, in distinct coordinate frames $A$ and $B$.
    Following (\ref{eq:residual}), the residual in frame $A$ (given some alignment $\pose[A]_B$) is
    \begin{equation}
      \vc[A]{r}_i = \vc[A]{p}_{a_i} - \pose[A]_B \circ \vc[B]{p}_{b_i}
    \end{equation}
    and the geometric compatibility GC is defined
    \begin{equation}\label{eq:gc}
      \dGC \triangleq \min_{\pose[A]_B \in \SE(3)} \vc{r}^T \big( \cov[A] + [ \rot[A]_B ] \cov[B] [ \rot[A]_B ]^T \big)^{-1} \vc{r}
    \end{equation}
    where $\vc{r}^T = [ \vc{r}_1^T, \vc{r}_2^T, \ldots, \vc{r}_m^T ]$ is the stacked residual vector (the frame superscripts are dropped for clarity), $\cov[A]$ and $\cov[B]$ are the $3m \times 3m$ covariance matrices in frames $A$ and $B$, respectively, and $[\rot] \triangleq \rm{BlockDiag}(\rot, \rot, \ldots, \rot) \in \SO(3m)$ is a $3m \times 3m$ block diagonal matrix.
    The resulting gating test $\dGC < \dchi{3m-6}$ is a \emph{posterior} compatibility test \cite{li2014fast} over $3m - 6$ degrees of freedom with the point estimates corresponding to $\CA$ and $\CB$ treated as independent sets of ``measurements.''
    This independence can be safely assumed if $\CA$ and $\CB$ are sufficiently separated in the graph, which can be enforced via \ref{uc:separation}.

    Note that evaluating GC involves a nonlinear optimization over $\SE(3)$.
    We achieve an efficient approximation in a two-step approach.
    First, we approximate (\ref{eq:gc}) with an orthogonal Procrustes optimization \cite{gower2004procrustes}
    \begin{equation}\label{eq:procrustes}
      \bar{\pose} = \argmin{\pose \in \SE(3)} \vc{r}^T \vc{r} = \argmin{\pose \in \SE(3)} \sum_{j=1}^m || \vc[A]{p}_{a_j} - T \circ \vc[B]{p}_{b_j} ||^2_2.
    \end{equation}
    whose solution $\bar{T}$ can be computed in closed-form.
    Then, we refine this estimate on the full objective function (\ref{eq:gc}) with a tangent-space linearization $\vc{r}_i = \vc[A]{r}_{a_i} - \big( \bar{T} \circ \rm{Exp}(\delta) \big) \circ \vc[B]{p}_{b_i}$ where $\delta \in \R^6$.
    By ``locking'' the covariance terms, this produces a small linear-least-squares problem over $\delta$ that can be solved efficiently.

    In contrast to the IPJC algorithm \cite{li2012ipjc}, our approach operates without a prior on $\pose[A]_B$, which would require access to $\cov[w]$.
    In practice, we find that our two-step optimization provides a suitable approximation of $(\ref{eq:gc})$ in constant-time.

  \subsection{Approximating Local Uncertainty}\label{ss:local_covs}
    \begin{figure}[t]
      % Tikz-augmented inkscape import
      \centering
      \begin{tikzpicture}[scale=0.8, every node/.style={transform shape}]
        \node[anchor=south west,inner sep=0] (image) at (0,0)
            {\includegraphics[width=\columnwidth]{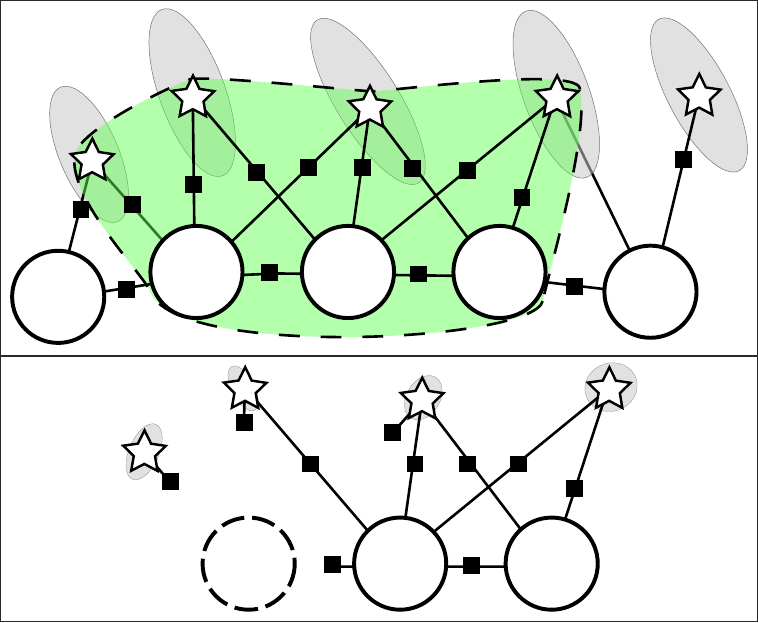}};
        % Add labels
        %\draw[help lines] (0,0) grid (9,7);

        \node at (0.85,6.83) {Frame $w$};
        \node at (0.4,5.2) {$\vc[w]{p}_1$};
        \node at (2.05,6.45) {$\vc[w]{p}_2$};
        \node at (4,6.4) {$\vc[w]{p}_3$};
        \node at (6.22,6.45) {$\vc[w]{p}_4$};
        \node at (7.73,6.5) {$\vc[w]{p}_5$};
        \node at (0.6,3.7) {$\pose[w]_0$};
        \node at (2.2,3.95) {$\pose[w]_1$};
        \node at (3.95,3.95) {$\pose[w]_2$};
        \node at (5.65,3.95) {$\pose[w]_3$};
        \node at (7.4,3.7) {$\pose[w]_4$};

        \node at (0.85,2.75) {Frame $\pose_1$};
        \node at (1.1,1.9) {$\vc[1]{p}_1$};
        \node at (2.3,2.6) {$\vc[1]{p}_2$};
        \node at (4.2,2.6) {$\vc[1]{p}_3$};
        \node at (6.35,2.65) {$\vc[1]{p}_4$};
        \node at (2.8,0.63) {$\pose[1]_1$};
        \node at (4.55,0.63) {$\pose[1]_2$};
        \node at (6.3,0.63) {$\pose[1]_3$};

      \end{tikzpicture}
      \caption{
          [top] Example SLAM graph expressed in the world frame.
                The green shaded region indicates the local subgraph extracted to approximate local uncertainty over $\vc{p}_3$.
          [bottom] This subgraph expressed in local frame of $\pose_1$.
                Note that marginal covariances (shaded ellipses) over the landmarks in this frame are generally smaller than in the world frame and show less correlation.
          }
      \label{fig:local_frames}
      \vspace{-0.5cm}
    \end{figure}
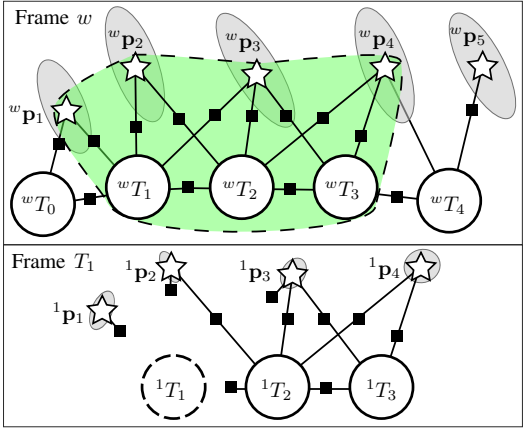

    In order to compute (\ref{eq:gc}), we need estimates of $\cov[A]$ and $\cov[B]$ in some to-be-determined frames $A$ and $B$.
    To simplify the following discussion, define $\mc{T}_j \subset \mc{T}$ to be the set of poses from which landmark $L_j$ is observed.
    Furthermore, assume that $\mc{T}_{\CA} \triangleq \bigcap_{i=1}^m \mc{T}_{a_i} \neq \emptyset$ and $\mc{T}_{\CB} \triangleq \bigcap_{i=1}^m \mc{T}_{b_i} \neq \emptyset$ -- that is that there exists at least one common pose adjacent to all of $\CA$ and another (distinct) pose adjacent to all of $\CB$.
    In the case that $\mc{T}_m$ are \emph{intervals}, this can be enforced efficiently during search with a pairwise locality constraint \ref{bc:locality}.
    Nevertheless, the approach outlined here can be straightforwardly extended to more general scenarios.

    One approach could be to select a $\pose_A \in \mc{T}_{\CA}$ and $\pose_B \in \mc{T}_{\CB}$ and compute $\cov[A]$ and $\cov[B]$ ``on-the-fly'' during search.
    However, because each landmark is eventually considered as part of numerous constellations, this can lead to significant redundant computation.
    Instead, we seek to pre-compute a set of \emph{independent} marginals for each $\vc{p}_j$, one for each local frame represented in $\mc{T}_j$.

    Our process is illustrated in Fig.~\ref{fig:local_frames} for a given $L_j$.
    We first extract the local subgraph containing $\vc{p}_j$, $\mc{T}_j$, and all landmarks adjacent to $\mc{T}_j$.
    For each $\pose_i \in \mc{T}_j$ we compute the $3 \times 3$ marginal $\cov[i]_jj$ over $\vc[i]{p}_j$ in the corresponding local frame.
    This computation involves only the local subgraph, and assuming a constant max cardinality $|\mc{T}_j| \leq N$, this can be accomplished in constant time for each $L_j$.
    Furthermore, because it involves only local information (and is independent of global linearization point), it can be performed incrementally (as only recent landmarks will need to be updated).
    Finally, because some information is ignored (specifically, observations of other local landmarks), this estimate is conservative.

    It should be noted that for nearby landmarks, the resulting estimates \emph{are} correlated, although in practice we've found these correlations to be small and the independence approximation to be sufficient in light of the computational advantages.
    In some scenarios (specifically pairwise comparisons), the effect of these correlations can be explicitly bounded, although for brevity further exploration is omitted here.

  \subsection{Linear-complexity GC}
    As stated before, we cache a set of marginals $\{ \pose[i]_{jj} \}$ for each landmark $\vc{p}_j$.
    During evaluation of (\ref{eq:gc}), the sets of common frames $\mc{T}_{\CA}$ and $\mc{T}_{\CB}$ are assured to be non-empty, and thus $\pose_A \in \mc{T}_{\CA}$ and $\pose_B \in \mc{T}_{\CB}$ can be selected via min-determinant or min-trace criteria.

    By assuming independence between each $\vc{p}_j$, $\cov[A]$ and $\cov[B]$ become block-diagonal, and (\ref{eq:gc}) simplifies to a linear sum-of-squares
    \begin{equation}\label{eq:gc_simp}
      \dGC = \sum_{i=j}^m \vc{r}_j^T (\cov[A]_{jj} + \rot[A]_B \cov[B]_{jj} \rot[A]_B^T)^{-1} \vc{r}_j
    \end{equation}
    Thus, the independence assumption allows linear-time evaluation of GC, compared to the generally quadratic evaluation of (\ref{eq:gc}).

%%%%%%% Sim setup images  %% include here so it can be top-aligned
\begin{figure}[t]
	\centering
	\begin{subfigure}[t]{.3\columnwidth}
		\centering
		\includegraphics[width=\linewidth]{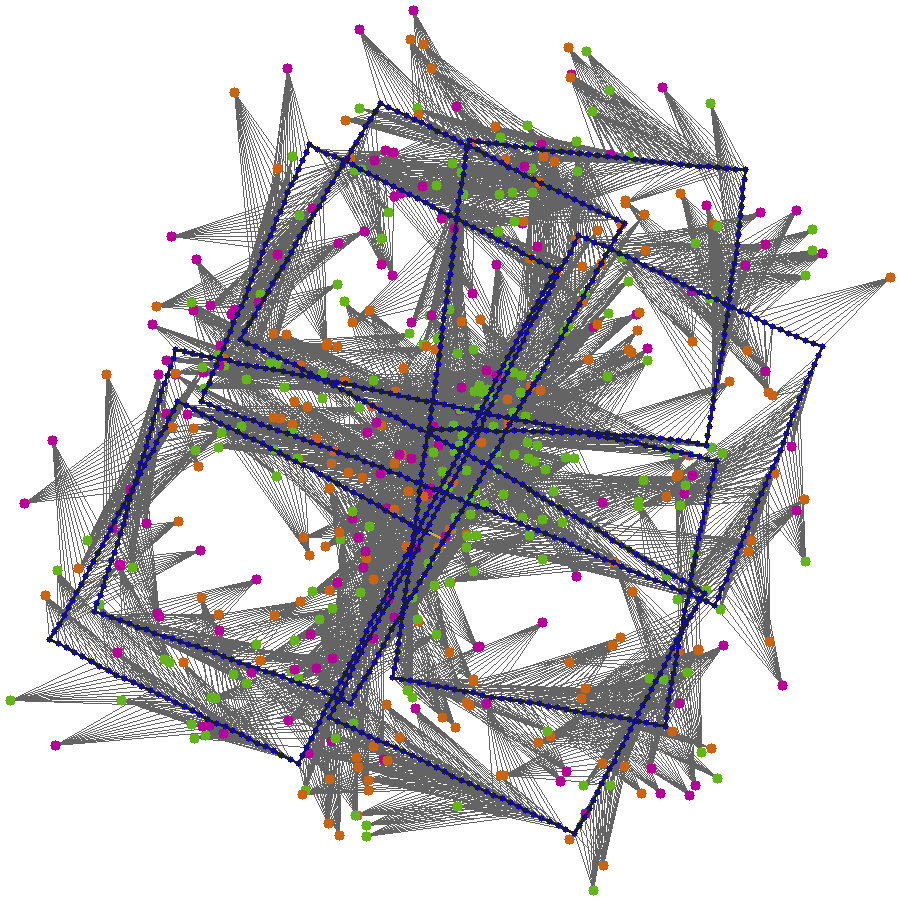}
		%\caption*{Full run w/o constellation merging.}
	\end{subfigure}
	\hfill
	\begin{subfigure}[t]{.3\columnwidth}
		\centering
		\includegraphics[width=\linewidth]{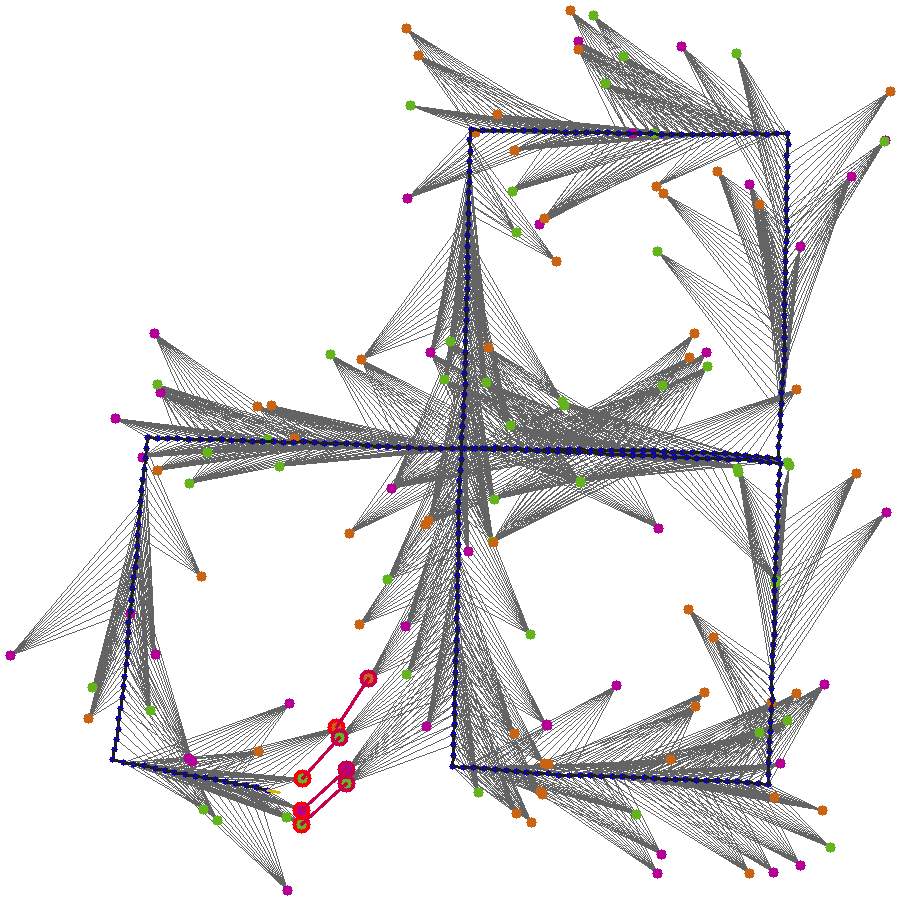}
		%\caption*{Early loop-closure with $m=4$.}
	\end{subfigure}
	\hfill
	\begin{subfigure}[t]{.3\columnwidth}
		\centering
		\includegraphics[width=\linewidth]{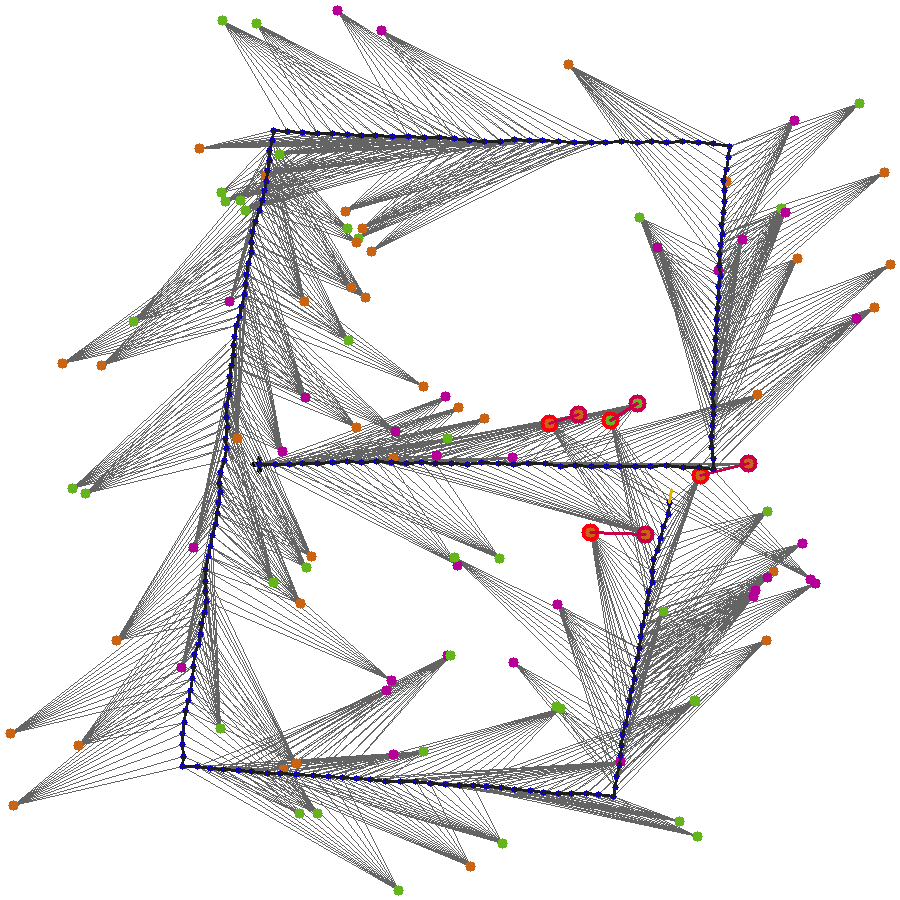}
		%\caption*{Later loop-closure after large rotation error.}
	\end{subfigure}
	\caption{Constellation merging in simulation. The robot ground-truth trajectory happens to be planar, but the state space is fully 3-dimensional.
		[left] Without constellation merging, localization error increases over time, and the final estimate shows significant error as well as many duplicate landmarks.
		[center, right] Correct constellation matches are detected despite significant drift in rotation and translation.}
	\label{fig:sim_setup}
	\vspace{-0.5cm}
\end{figure}

%%%%%%%%%%%%%%%%%%%%%%%%%%%%%%%%%%%%%%%%%%%%%%%
\section{Re-interpreting the Interpretation Tree}\label{s:hybrid_bb}
  Rather than defining hypotheses over an interpretation tree \cite{grimson1990object}, we explicitly approach the maximization (\ref{eq:jc_max}) as a set inclusion problem over $\Cliques \subset 2^{m \times n}$.
  This is represented as a binary search tree (BST), where each level of the BST corresponds to the inclusion or disclusion of a unary-feasible candidate $s_i \in \Vertices$.
  Every node in the tree (say at depth $d$) corresponds to a partial hypothesis $\C_d$ in which only $d \leq \lvert \Vertices \rvert$ candidates have been considered.
  Critically, inclusion of the next vertex $s_i$ is only considered if $s_i$ is adjacent to every previously included node $s_j$ in $\C_d$.
  This enforces that $\C_d \in \Cliques$ always, and can be implemented efficiently by maintaining a list of remaining unconsidered, but jointly-adjacent, nodes $S(\C_d) \subset \Vertices$ for each partial hypothesis $\C_d$.
  Given $C_d$ and $S(C_d)$, a tight upper bound is
  \begin{align}
    \forall \C \in \Cliques & : \C \supset \C_d \quad \text{we have} \nonumber \\
    \lvert \C \rvert &\leq \text{UpperBound}(\C_d) = \lvert \C_d \rvert + \lvert S(\C_d) \rvert.  \label{eq:upper_bound}
  \end{align}

  This BST approach is clearly more general than the interpretation tree, as it does \emph{not} implicitly enforce the constraint that each measurement corresponds to at most one landmark.
  However, this can be easily re-imposed via the binary disjointness constraint \ref{bc:disjoint}.
  When $\Cgraph$ reflects \emph{only} this disjointness constraint, the BST and interpretation tree approaches are identical.
  However, in the presence of other constraints, the BST is superior in that it only tests each candidate $s$ and pair of candidates $(s_i, s_j)$ \emph{once} (while building $\Cgraph$), and can leverage these constraints to provide a tighter upper bound via (\ref{eq:upper_bound}).
  In doing so, it unifies clique- and tree-based search schemes in a straightforward, easily-implemented way.

\vspace*{-.15cm}

%%%%%%%%%%%%%%%%%%%%%%%%%%%%%%%%%%%%%%%%%%%%%%
\section{Experimental Results}

\begin{figure}[t]
	% Tikz-augmented inkscape import
	\centering
	\begin{tikzpicture}[scale=0.94, every node/.style={transform shape}]
	\node[anchor=south west,inner sep=0] (image) at (0,0)
	{\includegraphics[width=\columnwidth,trim={3.5cm, 8.6cm, 3.3cm, 8.4cm},clip]{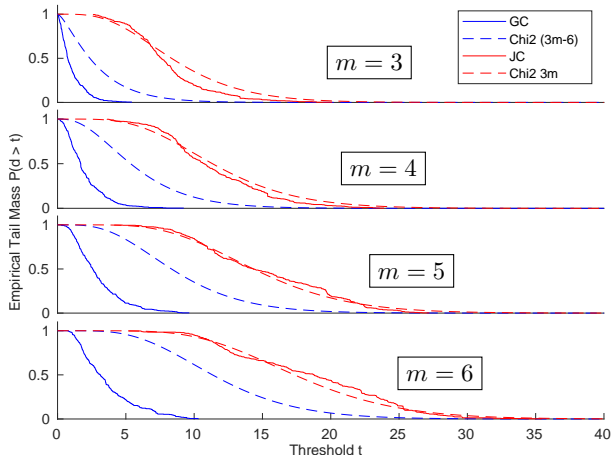}};
	% Add labels
	%\draw[help lines] (0,0) grid (9,8);

	\node[draw] at (5.1,5.55) {$m = 3$};
	\node[draw] at (5.3,4.05) {$m = 4$};
	\node[draw] at (5.7,2.6) {$m = 5$};
	\node[draw] at (6.1,1.15) {$m = 6$};

	\end{tikzpicture}
	\vspace*{-0.1cm}
	\caption{Empirical compatibility histograms (solid) vs.\ corresponding $\chi^2$ tail mass (dashed) for true constellation matches of varying cardinalities.
    For the most part the tail mass of the JC metric follows the prediction, but shows some discrepancy due to global-frame nonlinearity.
		As expected, our conservative GC estimates decay faster than the corresponding $\chi^2$ tail, ensuring that we do not ``miss'' good matches.
	}
	\label{fig:histograms}
	\vspace*{-0.1cm}
\end{figure}

  The proposed statistics and methods were validated in a simulated nonlinear visual-SLAM setting implemented with GTSAM \cite{dellaert2012factor}.
  As the robot moves along a loopy trajectory shown in Fig.~\ref{fig:sim_setup}, it receives noisy odometry and makes observations of randomly-distributed landmarks via a simulated single-camera sensor, with limited range and field-of-view.
  To simulate a ``short-term'' data association solution (e.g.\ frame-to-frame tracking), landmark associations are ``lost'' once the feature leaves the camera field of view, and further observations are assigned to a new, duplicate landmark.
  Thus, if no map-merging is performed, there is no global loop-closure, and localization uncertainty grows with time.

  To emulate semantic SLAM, landmarks are randomly assigned one of three classes (indicated by color), and it is assumed that class label is accurately observed (reasonable given the performance of state-of-the-art detectors \cite{redmon2016you}).
  This semantic information is used as a unary constraint \ref{uc:class} to help sparsify the correspondence graph.

\begin{figure}[t!]
	\centering
	\begin{tikzpicture}[scale=0.94, every node/.style={transform shape}]
	\node[anchor=south west,inner sep=0] (image) at (0,0)
	{%\includegraphics[width=\linewidth,trim={3.3cm, 9.35cm, 3.5cm, 9.15cm},clip]{sim_incr_results}
	 \includegraphics[width=\linewidth,trim={3.3cm, 8.35cm, 3.5cm, 8.0cm},clip]{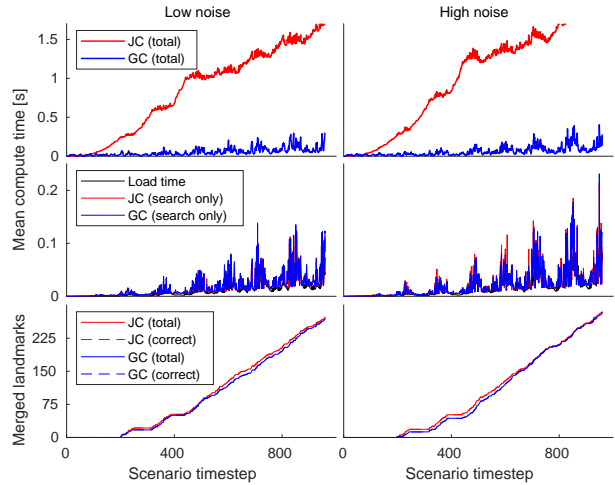}};
  \end{tikzpicture}
	\vspace*{-0.15cm}
	\caption{Averaged simulation results over low and high noise conditions, comparing joint-compatibility to our method.
		Both methods identify comparable numbers of merges (bottom), and achieve over $99\%$ accuracy in all tests.
    The $\cov[w]$ pre-computation required by JC dominates total computation times (top), while the time spent in actual tree search (middle) is similar for both methods.
    Note the difference in axis scales.
		The time spent updating the correspondence graph $\Cgraph$ (common to both methods) is shown in black.
	}
	\label{fig:sim_incr_results}
	\vspace{-0.1cm}
\end{figure}

  Fig.~\ref{fig:histograms} demonstrates the statistical consistency of our GC metric on randomly-sampled \emph{ground-truth} constellation matches in simulation.
  Because of lossy linearizing approximations, the JC metric does not perfectly follow a $\chi^2$ distribution, whereas the GC scores are (correctly) conservative.
  This conservatism arises from the practical need to estimate landmark uncertainties using only a limited subset of the available data (see Sec.~ \ref{ss:local_covs}).
  Here we use a minimal subset, although larger subsets could be chosen (at the cost of more computation).
  Fig.~\ref{fig:sim_incr_results} shows a comparison in both detection performance and computation time between the proposed GC metric and baseline JC \cite{neira2001data}.
  As can be seen, a similar number of matches are found using the GC, but without the expensive step of computing $\cov[w]$.
  Thus, our GC method performs as well as a full JC-based search but at a fraction of the computational cost.

%\vspace*{-.06cm}

%%%%%%%%%%%%%%%%%%%%%%%%%%%%%%%%%%%%%%%%%%%%
\section{CONCLUSIONS}
  Measurements are most informative when the estimate has drifted, but that is when they are simultaneously the most ambiguous.
  Given the catastrophic risks of incorrect associations, it is always safer to ascribe ambiguous measurements to a new landmark than to an existing one.
  With this principle in mind, this paper introduces an efficient method of ``delayed'' data association via landmark constellation merging.
  While most relevant methods assume access to the full covariance matrix and/or fully uncorrelated measurements (i.e.\ in feature cloud matching), our method leverages local and incrementally-computable information to identify good candidates over the full graph.
  If needed, the sparse set of resulting matches can then be verified via standard covariance-based methods at a computational cost that is tenable in practice.
  We believe that our GC-based approach provides a robust, secondary level of loop-closure detection \emph{in the back-end} that facilitates the re-capture of ``missed'' loop closures, reducing the burden on front-end data association.

%%%%%%%%%%%%%%%%%%%%%%%%%%%%%%%%%%%%%%%%%%%%%%%%%%%%%%%%%%%%%%%%%%%%%%%%%%%%%%%%
%\section*{APPENDIX}


%\addtolength{\textheight}{-3cm}   % This command serves to balance the column lengths

%\vspace*{-.06cm}

\section*{ACKNOWLEDGMENT}
  Thanks to Dr.\ Kasra Khosoussi for the many productive conversations and input over the course of this work.

%%%%%%%%%%%%%%%%%%%%%%%%%%%%%%%%%%%%%%%%%%%%%%%%%%%%%%%%%%%%%%%%%%%%%%%%%%%%%%%%
\balance

\bibliographystyle{IEEEtran}
\bibliography{IEEEabrv,ref}

\end{document}